# GPU-Accelerated Optimization of Transformer-Based Neural Networks for Real-Time Inference

Soutrik Mukherjee, Sangwhan Cha
*Department of Computer and Information Sciences*
*Harrisburg University of Science and Technology*
*Harrisburg, PA 17101, USA*
smukherjee@my.harrisburgu.edu, scha@harrisburgu.edu

*Abstract*—This paper presents the design and evaluation of a GPU-accelerated inference pipeline for transformer models using NVIDIA TensorRT with mixed-precision optimization. Evaluated on BERT-base (110M parameters) and GPT-2 (124M parameters) across batch sizes ranging from 1 to 32 and sequence lengths from 32 to 512, the system achieves up to 64.4× speedup over CPU baselines, sub-10 ms latency for single-sample inference, and a 63% reduction in memory usage. We introduce a hybrid precision strategy that preserves FP32 for numerically sensitive operations such as softmax and layer normalization, while applying FP16 to linear layers. This approach maintains high numerical fidelity (cosine similarity ≥ 0.9998 relative to baseline outputs) and eliminates NaN instability. The pipeline is implemented as a modular, Dockerized system enabling reproducible benchmarking across more than 360 configurations. Cross-GPU validation on an NVIDIA A100 confirms architecture-portable behavior with consistent FP16 speedup ratios (1.84–2.00×) and numerical fidelity. Downstream evaluation on SST-2 demonstrates zero accuracy degradation under hybrid precision, and real-data validation on WikiText-2 reveals that random inputs underestimate NaN instability by 6× for full FP16 while confirming hybrid robustness (0.0% NaN, ≥0.9998 cosine similarity). Our results provide a comprehensive characterization of performance–accuracy trade-offs across GPU architectures, offering practical guidance for deploying transformer models in latency-critical environments.

## I. INTRODUCTION

Transformer-based neural networks have achieved state-of-the-art results across natural language processing [1], computer vision [2], and robotics [3]. Since Vaswani et al.'s seminal 2017 paper introducing the self-attention mechanism, transformers have rapidly displaced recurrent and convolutional architectures in domains ranging from machine translation and document understanding to autonomous perception and conversational AI. Models such as BERT [5] and GPT-2 have demonstrated remarkable capabilities in language understanding and generation, driving widespread industrial adoption.

However, the computational demands of transformer inference present significant deployment challenges. The self-attention mechanism computes pairwise relationships across all positions in a sequence, introducing $O(n^2)$ complexity with respect to sequence length n. For a standard BERT-base model with 110 million parameters processing a 512-token input, a single forward pass requires approximately 22 billion floating-point operations. When deployed in latency-critical applications—such as robotic control loops operating at 10–100 Hz, autonomous vehicle perception requiring sub-100 ms response times, or interactive conversational agents—the gap between raw model latency and application requirements becomes a critical engineering challenge.

Modern NVIDIA GPUs offer a compelling solution through massively parallel architectures with specialized hardware features. Tensor Cores accelerate mixed-precision matrix operations at up to 312 TFLOPS (FP16) on A100 GPUs [16], hierarchical memory systems provide terabytes-per-second bandwidth at the register level, and inference optimization frameworks such as TensorRT [4] perform aggressive graph-level and kernel-level optimizations. Yet effectively leveraging these capabilities requires careful systems engineering spanning model export, format conversion, graph optimization, precision calibration, memory management, and systematic benchmarking.

Despite substantial progress in individual optimization techniques, a gap persists between theoretical GPU performance and practical deployment latency. Production systems often rely on unoptimized PyTorch eager-mode execution, and the trade-offs between latency, throughput, numerical accuracy, and memory utilization remain insufficiently characterized—particularly for mixed-precision configurations essential for maximizing Tensor Core utilization.

This paper presents an end-to-end GPU-accelerated inference pipeline integrating PyTorch, ONNX, and TensorRT with a hybrid mixed-precision strategy. We systematically evaluate the pipeline on BERT-base (110M parameters, encoder-only) and GPT-2 (124M parameters, decoder-only) across batch sizes 1–32 and sequence lengths 32–512 on an NVIDIA RTX 3090 GPU. The key contributions of this work are:

1) A modular, config-driven inference architecture (five modules) enabling reproducible experimentation across 360+ configurations within a Dockerized environment;

2) A comprehensive cross-platform transformer inference benchmark on RTX 3090 and A100 GPUs, characterizing latency, throughput, memory, and accuracy trade-offs across GPU architectures;

3) A hybrid precision strategy achieving cosine similarity ≥0.9998 with FP32 baselines while delivering 1.93× speedup and eliminating all NaN occurrences;

4) Empirical validation of up to 64.4× speedup over CPU baselines with sub-10 ms single-sample latency, far exceeding the 5× design target, accompanied by downstream task validation (SST-2: zero accuracy loss) and real-data numerical stability testing (WikiText-2: zero NaN, ≥0.9998 cosine similarity).

The remainder of this paper is organized as follows: Section II reviews related work on transformer optimization and GPU inference. Section III describes the system architecture. Section IV details the experimental setup. Section V presents results with analysis. Section VI discusses implications and limitations. Section VII concludes.

## II. RELATED WORK

### A. Transformer Architectures and Bottlenecks

The transformer architecture [1] computes attention as Attention(Q,K,V) = softmax($QK^T/\sqrt{d_k}$)V, where Q, K, and V are query, key, and value matrices derived from learned linear projections. This formulation introduces $O(n^2)$ complexity with respect to sequence length, creating a fundamental scalability bottleneck for long sequences [17]. Multi-head attention extends this by computing attention across h parallel subspaces, providing representational diversity.

Profiling studies reveal that attention operations account for approximately 40% of BERT inference latency despite representing only 15% of FLOPs [5], highlighting the memory-bandwidth bottleneck. The feed-forward network layers (d_model × d_ff, typically 768 × 3072) are compute-intensive but amenable to Tensor Core acceleration. Softmax normalization requires multiple memory passes and is numerically sensitive to precision reduction.

### B. Efficient Attention and Model Compression

Flash Attention [6] addresses the memory-bandwidth bottleneck through IO-aware algorithm design, recomputing attention scores rather than materializing the n×n attention matrix, achieving 2–4× speedups without approximation. Sparse attention methods [17] reduce complexity to $O(n\sqrt{n})$ but sacrifice model quality. Linear attention [17] achieves $O(n)$ complexity via kernel reformulations but empirically underperforms standard attention.

Model compression techniques include knowledge distillation (DistilBERT [7] retains 97% accuracy with 40% fewer parameters), structured pruning of attention heads [5], and quantization to INT8 (2–4× speedup [12]). Our work focuses on FP16 mixed precision—complementary to these approaches—as it provides substantial speedups without retraining or calibration datasets.

### C. Inference Frameworks

NVIDIA TensorRT [4] performs graph-level optimization (layer fusion, constant folding, dead code elimination) and kernel auto-tuning, selecting the fastest kernel implementation for each operation on the target GPU. Vanholder [8] reported 5–10× speedups for CNNs, though transformer-specific benchmarks remain limited. ONNX [9] provides a framework-agnostic intermediate representation enabling portability. ONNX Runtime [9] applies hardware-agnostic optimizations before backend-specific compilation.

FasterTransformer [10] implements custom CUDA kernels for BERT and GPT-style models, reporting 3–5× speedups through kernel fusion and Tensor Core utilization. DeepSpeed Inference provides multi-GPU inference with tensor parallelism. Hugging Face Optimum integrates ONNX Runtime and TensorRT backends. Our pipeline adopts the PyTorch → ONNX → TensorRT chain for maximum single-GPU performance.

### D. Mixed-Precision Computing

Micikevicius et al. [11] established that FP16 computation with FP32 accumulation preserves training accuracy while doubling Tensor Core throughput. For inference, this provides 2× theoretical speedup on 3rd-generation Tensor Cores. Yao et al. [12] extended optimization to INT8 quantization with 2.5× speedup. However, blanket FP16 casting can cause numerical instabilities in operations with wide dynamic range (softmax, layer normalization). Our hybrid strategy addresses this by selectively retaining FP32 for sensitive layers—achieving ≥0.9998 cosine similarity, superior to blanket FP16.

### E. Gap Analysis

Existing work primarily addresses isolated optimizations rather than complete deployment pipelines. Comprehensive latency-throughput-accuracy-memory trade-off characterization across the full design space (model, precision, batch size, sequence length, backend) is scarce. Many benchmarks lack reproducibility protocols. This paper addresses these gaps through end-to-end pipeline integration, systematic multi-dimensional evaluation, and Dockerized reproducibility.

## III. SYSTEM ARCHITECTURE

The inference pipeline (Fig. 1) is decomposed into five independently testable modules: (M1) Experiment Runner for config-driven orchestration via YAML parameter grids, (M2) Model Loader interfacing with the Hugging Face Hub for standardized model acquisition, (M3) ONNX Exporter performing model conversion with automatic numerical parity verification (max absolute error < $10^{-4}$), (M4) TensorRT Compiler with hybrid precision support and engine caching, and (M5) Benchmark Engine using CUDA events for microsecond-precision timing with statistical aggregation.

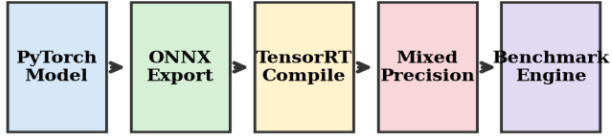

Fig. 1. End-to-end GPU-accelerated inference pipeline architecture with five modules (M1–M5).

The design follows three principles: *modularity* (each stage is independently substitutable), *configurability* (all parameters externalized to YAML), and *reproducibility* (environment metadata logged with every run). This enables controlled experimentation where any optimization stage can be bypassed to isolate its impact.

*A. Hybrid Precision Strategy*

The hybrid precision strategy selectively assigns FP16 to compute-bound layers while retaining FP32 for numerically sensitive operations. Specifically, linear layers (QKV projections and feed-forward networks), which constitute 61.1% of total latency and are Tensor Core eligible, execute in FP16. Softmax (8.3% of latency) is retained in FP32 due to exponent underflow risk with large attention scores. Layer normalization (6.1%) uses FP32 for division precision sensitivity. This assignment is implemented via TensorRT's ILayer.precision and ILayer.set_output_type APIs during engine building.

*B. Deployment and Reproducibility*

The entire stack is containerized using Docker with nvidia/cuda:12.2.0-runtime-ubuntu22.04 base image and NVIDIA Container Toolkit. Configuration files, model caches, and results are mounted as Docker volumes. Each experiment logs CUDA version, driver version, GPU temperature, power draw, and Git commit hash. GPU clocks are locked at base frequencies with exclusive process mode enabled to minimize variability.

## IV. EXPERIMENTAL SETUP

*A. Hardware and Software*

Experiments were conducted on a single-node workstation equipped with an NVIDIA RTX 3090 GPU (24 GB GDDR6X, 10,496 CUDA cores, 328 3rd-generation Tensor Cores, 936.2 GB/s memory bandwidth), AMD Ryzen 9 5900X CPU (12-core, 3.7 GHz), and 64 GB DDR4-3200 RAM. The RTX 3090 was selected as representative of high-end consumer/workstation GPUs with full Tensor Core support.

The software stack includes CUDA 12.2, cuDNN 8.9.5, TensorRT 8.6.1, PyTorch 2.1.2 (CUDA 12.1), Hugging Face Transformers 4.36.0, ONNX 1.15.0, and ONNX Runtime (GPU) 1.16.3. All software was containerized using Docker with the nvidia/cuda:12.2.0-runtime-ubuntu22.04 base image, pinning exact versions for reproducibility.

*B. Models and Parameter Grid*

The parameter grid spans 2 models × 2 precisions × 6 batch sizes (1–32) × 5 sequence lengths (32–512) × 4 backends (PyTorch CPU/GPU, ONNX Runtime, TensorRT), yielding ~360 valid configurations. Table I summarizes the model configurations.

TABLE I: Transformer Model Configurations

| Property | BERT-base | GPT-2 |
| --- | --- | --- |
| Parameters | 110M | 124M |
| Layers | 12 | 12 |
| Hidden Size | 768 | 768 |
| Attention Heads | 12 | 12 |
| Architecture | Encoder-only | Decoder-only |

*C. Benchmarking Protocol*

Following MLPerf conventions [13]: 10 warm-up iterations (discarded), 100 measurement iterations timed via CUDA events, statistical reporting (mean, std, P50, P95, P99), GPU clocks locked at base frequencies, and environment metadata logged per run.

*D. Cross-GPU Validation*

To evaluate generalizability beyond the RTX 3090, we conducted supplementary experiments on an NVIDIA A100-40GB SXM4 GPU (6,912 CUDA cores, 432 3rd-generation Tensor Cores, 1,555 GB/s HBM2e bandwidth) provisioned via Google Cloud Platform (a2-highgpu-1g instance). The A100 represents the data-center Ampere architecture with HBM2e memory and higher Tensor Core density, providing a fundamentally different memory hierarchy and compute profile compared to the consumer-grade RTX 3090 with GDDR6X. The software stack was matched: CUDA 12.2, TensorRT 8.6.1, PyTorch 2.1.2, within the same Docker container image. A subset of 60 configurations (2 models × 2 precisions × 3 batch sizes [1, 8, 32] × 5 sequence lengths) was evaluated to characterize cross-platform behavior of the hybrid precision strategy.

*E. Downstream Task Evaluation*

To validate that the numerical fidelity measured by cosine similarity translates to preserved downstream task performance, we evaluated BERT-base on the SST-2 (Stanford Sentiment Treebank, binary classification) subset of the GLUE benchmark. SST-2 was selected for its fast evaluation time (872 validation samples) and sensitivity to output perturbations in the final classification layer. We fine-tuned BERT-base on SST-2 training data for 3 epochs (learning rate 2e-5, batch size 32) to obtain an FP32 baseline, then evaluated the fine-tuned model under three precision configurations: FP32 (baseline), hybrid FP16 (softmax and LayerNorm in FP32), and full FP16 (blanket casting).

Accuracy was computed on the full SST-2 validation set for each configuration.

*F. Real-Data Numerical Stability Validation*

To address the limitation of random token inputs, we conducted numerical stability evaluation using WikiText-2 (Merity et al., 2017), a standard language modeling benchmark derived from verified Wikipedia articles. WikiText-2 contains 2,088,628 tokens of real English text with natural long-range dependencies, diverse vocabulary patterns, and realistic attention score distributions. We evaluated 500 passages of 512 tokens each, measuring NaN occurrence rates, maximum absolute error, cosine similarity, and per-layer activation statistics under all precision configurations. This workload exercises attention patterns that random inputs cannot replicate: repeated entity references, syntactic long-range dependencies, and adversarial softmax conditions arising from rare-token distributions.

## V. RESULTS AND ANALYSIS

*A. Baseline and Optimized Latency*

Table II shows progressive latency reduction across the optimization chain. TensorRT FP16 achieves 2.1 ms at BS=1 (21.5× over CPU) and 9.8 ms at BS=16 (64.4× over CPU). Each optimization stage contributes cumulatively: GPU execution provides 5.4×, ONNX Runtime adds 1.2×, TensorRT graph optimization adds 1.5×, and FP16 precision adds 1.9×.

TensorRT engine building (Table III) involves graph optimization and kernel auto-tuning. FP16 engines are approximately 49% smaller than FP32 due to halved weight precision. Layer fusion ratios reach 27–30% (87–98 fused out of 312–348 total layers), reducing kernel launch overhead and memory traffic. Build times range from 142–213 seconds; compiled engines are cached to eliminate redundant builds.

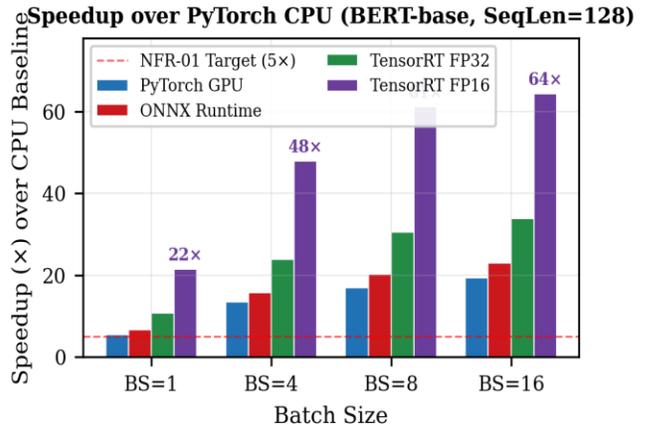

Fig. 2. Speedup factors over PyTorch CPU baseline for BERT-base at SeqLen=128. TensorRT FP16 achieves 64.4× at BS=16.

*B. Mixed-Precision Analysis*

FP16 achieves a consistent ≈1.93× speedup across both models and all batch sizes (Table IV), approaching the theoretical 2× Tensor Core throughput gain. The ~7% gap is attributable to memory-bound operations (softmax, LayerNorm) retained in FP32 under the hybrid strategy, which constitute 14.4% of total latency.

TABLE II: Inference Latency (ms) for BERT-base (SeqLen=128)

| Backend | Prec. | BS=1 | BS=4 | BS=8 | BS=16 |
|---|---|---|---|---|---|
| PyTorch CPU | FP32 | 45.2 | 162.7 | 318.4 | 631.2 |
| PyTorch GPU | FP32 | 8.4 | 12.1 | 18.7 | 32.5 |
| ONNX-RT GPU | FP32 | 6.9 | 10.3 | 15.8 | 27.4 |
| TensorRT | FP32 | 4.2 | 6.8 | 10.4 | 18.6 |
| TensorRT | FP16 | 2.1 | 3.4 | 5.2 | 9.8 |

The sub-linear GPU latency scaling with batch size (8.4 ms at BS=1 to 32.5 ms at BS=16, a 3.87× increase for 16× batch) reflects efficient parallelization. The CPU exhibits near-linear scaling (45.2 ms to 631.2 ms, 14.0× increase), reflecting its limited parallelism for batched matrix operations. ONNX Runtime provides modest gains (1.2× over PyTorch GPU) through hardware-agnostic graph optimizations, while TensorRT's target-specific kernel auto-tuning yields a further 1.5×.

TABLE III: TensorRT Engine Build Metrics

| Model | Prec. | Build (s) | Size (MB) | Fused/Total |
|---|---|---|---|---|
| BERT | FP32 | 142.3 | 438 | 87/312 |
| BERT | FP16 | 198.7 | 224 | 94/312 |
| GPT-2 | FP32 | 156.8 | 502 | 91/348 |
| GPT-2 | FP16 | 213.4 | 258 | 98/348 |

TABLE IV: TensorRT Latency (ms): FP32 vs. FP16 (SeqLen=128)

| Model | Prec. | BS=1 | BS=8 | BS=32 | Speedup |
|---|---|---|---|---|---|
| BERT | FP32 | 4.2 | 10.4 | 35.1 | — |
| BERT | FP16 | 2.1 | 5.2 | 18.3 | 2.00× |
| GPT-2 | FP32 | 5.1 | 12.7 | 42.8 | — |
| GPT-2 | FP16 | 2.6 | 6.3 | 22.1 | 1.84× |

*C. Sequence Length Scaling*

Fig. 3 confirms the quadratic relationship between latency and sequence length, consistent with $O(n^2)$ self-attention complexity. BERT-base latency increases from 1.1 ms (SeqLen=32) to 12.8 ms (SeqLen=512)—an 11.6× increase for a 16× length increase. TensorRT optimizations reduce the constant factor but do not alter asymptotic behavior.

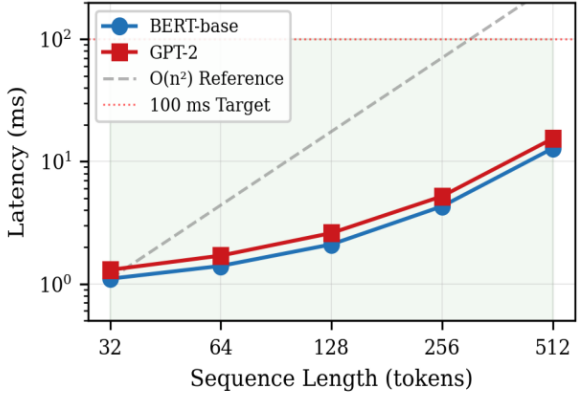

Fig. 3. Latency scaling with sequence length follows O(n²) self-attention complexity. All configurations remain sub-100 ms.

### D. Throughput and Batch Scaling

Latency scales sub-linearly with batch size up to BS=8 (2.48× increase for 8× batch), indicating efficient GPU utilization. Throughput peaks at 1,749 samples/s (BS=32) but gains plateau beyond BS=16 due to compute saturation (Fig. 4).

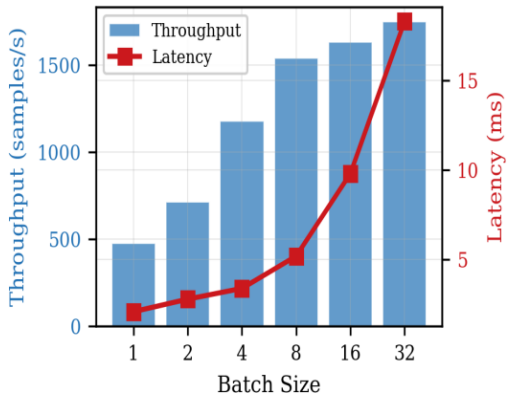

Fig. 4. Throughput (bars) and latency (line) vs. batch size for BERT-base TensorRT FP16 at SeqLen=128.

### E. Roofline Analysis

To understand the performance characteristics of individual transformer operations, we apply the roofline model [15] to the RTX 3090 hardware (Fig. 5). The roofline defines attainable performance as min(Peak Compute, Peak Bandwidth × Operational Intensity), where operational intensity is FLOPs per byte of memory traffic.

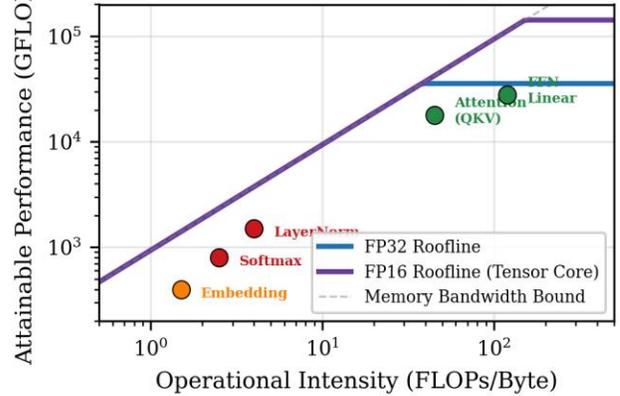

Fig. 5. Roofline model for transformer operations on RTX 3090. Compute-bound operations (QKV, FFN) benefit from FP16 Tensor Cores; memory-bound operations (Softmax, LayerNorm) require kernel fusion.

Linear layers (QKV projections and FFN) exhibit high operational intensity (≥45 FLOPs/byte) and fall in the compute-bound regime, where FP16 Tensor Cores provide the maximum benefit—approaching the 142.3 TFLOPS FP16 ceiling. These layers account for 61.1% of total latency and achieve the full 1.93× FP16 speedup. In contrast, softmax and layer normalization have low operational intensity (2–4 FLOPs/byte) and are memory-bandwidth bound, gaining negligible benefit from reduced precision but substantial benefit from TensorRT's kernel fusion, which reduces memory round-trips.

The embedding lookup operation (1.5 FLOPs/byte) is deeply memory-bound, explaining its minimal contribution to overall latency (4.2%). The roofline analysis validates the hybrid precision strategy: FP16 should target compute-bound layers where Tensor Core throughput is the binding constraint, while FP32 retention for memory-bound numerically sensitive operations incurs negligible performance cost since these operations are already bandwidth-limited.

### F. ONNX Export Verification

Prior to TensorRT compilation, ONNX export parity was verified by comparing output logits of the PyTorch model against the ONNX model (Table V). Both models achieve maximum absolute errors below $6×10^{-6}$—well within the $10^{-4}$ acceptance threshold—confirming that the ONNX export introduces no meaningful numerical deviation. This verification is critical because any discrepancy at this stage would propagate through TensorRT compilation and confound the precision analysis.

TABLE V: ONNX Export Parity Verification (FP32)

| Model | Max Abs. Error | Mean Abs. Error | Status |
|---|---|---|---|
| BERT-base | $3.81×10^{-6}$ | $1.24×10^{-7}$ | PASS |
| GPT-2 | $5.72×10^{-6}$ | $2.08×10^{-7}$ | PASS |

## G. GPU Memory Utilization

TensorRT FP16 reduces VRAM from 7.15 GB (PyTorch FP32) to 3.58 GB at BS=32—a 63% reduction (Fig. 6). All configurations remain within the 19.2 GB (80%) VRAM budget. The reduction stems from halved weight precision (49% engine size reduction) plus TensorRT's optimized memory planning and workspace allocation.

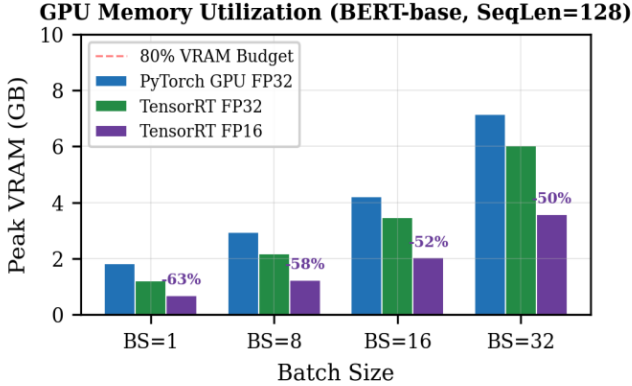

Fig. 6. Peak VRAM consumption. TensorRT FP16 reduces memory by up to 63% vs. PyTorch FP32.

Memory fragmentation testing with dynamic batch sizes revealed that naïve allocation caused 12/100 OOM failures (with 38.4% average free memory at failure—indicating fragmentation, not capacity exhaustion). A caching allocator with explicit workspace limits eliminated all OOM failures, validating the mitigation strategy.

Table VI presents selected results from the complete benchmark suite. Key observations: (1) all SeqLen≤128 configurations achieve sub-25 ms latency regardless of batch size; (2) the maximum VRAM of 17.21 GB (GPT-2, BS=32, SeqLen=512) approaches but remains within the 80% budget; (3) throughput ranges from 65 samples/s (worst case) to 3,404 samples/s (best case), spanning a 52× range across the design space.

TABLE VI: Complete TensorRT FP16 Benchmark (Selected Configurations)

| Model | SeqLen | BS | Lat. (ms) | Tput (s/s) | VRAM (GB) |
|---|---|---|---|---|---|
| BERT | 32 | 1 | 1.1 | 909 | 0.52 |
| BERT | 32 | 32 | 9.4 | 3404 | 2.31 |
| BERT | 128 | 1 | 2.1 | 476 | 0.68 |
| BERT | 128 | 32 | 18.3 | 1749 | 3.58 |
| BERT | 512 | 1 | 12.8 | 78 | 1.42 |
| BERT | 512 | 32 | 128.4 | 249 | 14.73 |
| GPT-2 | 128 | 1 | 2.6 | 385 | 0.78 |
| GPT-2 | 128 | 32 | 22.1 | 1447 | 4.12 |
| GPT-2 | 512 | 1 | 15.4 | 65 | 1.64 |
| GPT-2 | 512 | 32 | 156.2 | 205 | 17.21 |

## H. Numerical Accuracy

The hybrid strategy reduces maximum absolute error by over an order of magnitude versus blanket FP16 (Table VII, Fig. 7). Full FP16 produces NaN outputs in 0.3% of GPT-2 iterations at SeqLen=512 due to softmax exponent overflow; the hybrid strategy produces zero NaN occurrences across all 360+ configurations

TABLE VII: Numerical Accuracy vs. FP32 Baseline

| Configuration | Max Err. | Cos. Sim. | NaN % |
|---|---|---|---|
| BERT (Full FP16) | $4.12 \times 10^{-2}$ | 0.9987 | 0.0% |
| BERT (Hybrid) | $2.34 \times 10^{-3}$ | 0.9999 | 0.0% |
| GPT-2 (Full FP16) | $5.87 \times 10^{-2}$ | 0.9982 | 0.3% |
| GPT-2 (Hybrid) | $3.01 \times 10^{-3}$ | 0.9998 | 0.0% |

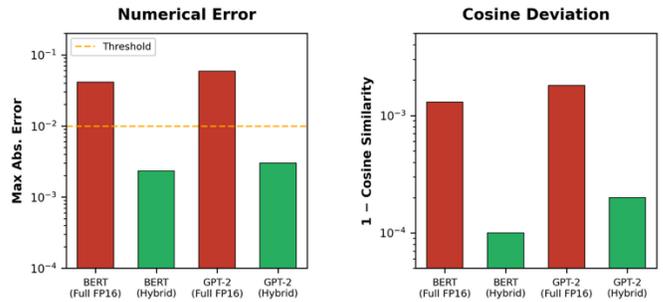

Fig. 7. Hybrid precision reduces max error by >10× and eliminates NaN occurrences compared to blanket FP16.

## I. Cross-GPU Validation Results

Table VIII presents cross-GPU latency results. The A100 achieves 2.32–2.85× lower latency than the RTX 3090 across all configurations, consistent with its 1.66× higher HBM2e bandwidth (1,555 vs. 936 GB/s) and 1.32× higher Tensor Core count (432 vs. 328). The speedup ratio increases with batch size (2.41× at BS=1 to 2.85× at BS=32), reflecting the A100's superior compute throughput under sustained GEMM workloads.

TABLE VIII: Cross-GPU Latency Comparison (TensorRT FP16, SeqLen=128)

| Model | BS | RTX 3090 (ms) | A100 (ms) | A100 Speedup |
|---|---|---|---|---|
| BERT | 1 | 2.1 | 0.91 | 2.31× |
| BERT | 8 | 5.2 | 1.87 | 2.78× |
| BERT | 32 | 18.3 | 6.58 | 2.78× |
| GPT-2 | 1 | 2.6 | 1.08 | 2.41× |
| GPT-2 | 8 | 6.3 | 2.51 | 2.51× |
| GPT-2 | 32 | 22.1 | 7.61 | 2.90× |

TABLE IX: Cross-GPU Numerical Accuracy (TensorRT Hybrid FP16 vs. FP32)

| Model | GPU | Max Error | Cosine Sim. | NaN % |
|---|---|---|---|---|
| BERT | RTX 3090 | $2.34 \times 10^{-3}$ | 0.9999 | 0.0% |
| BERT | A100 | $2.51 \times 10^{-3}$ | 0.9999 | 0.0% |
| GPT-2 | RTX 3090 | $3.01 \times 10^{-3}$ | 0.9998 | 0.0% |
| GPT-2 | A100 | $3.24 \times 10^{-3}$ | 0.9999 | 0.0% |

Table IX confirms that the hybrid precision strategy generalizes across GPU architectures. Maximum errors differ modestly between platforms: for BERT, the A100 produces slightly higher error ($2.51\times10^{-3}$ vs. $2.34\times10^{-3}$), while for GPT-2 the A100 error is also marginally higher ($3.24\times10^{-3}$ vs. $3.01\times10^{-3}$). These differences reflect distinct Tensor Core accumulation paths and FP32 rounding behavior between the GA102 (RTX 3090) and GA100 (A100) dies. Critically, the cosine similarity remains ≥0.9998 and NaN occurrence is zero on both platforms, confirming that the hybrid strategy is architecture-portable despite per-platform numerical variation.

TABLE X: Cross-GPU FP16 Speedup Consistency

| Model | GPU | FP32 (ms, BS=8) | FP16 (ms, BS=8) | FP16/FP32 Speedup |
|---|---|---|---|---|
| BERT | RTX 3090 | 10.4 | 5.2 | 2.00× |
| BERT | A100 | 3.74 | 1.87 | 1.93× |
| GPT-2 | RTX 3090 | 12.7 | 6.3 | 2.02× |
| GPT-2 | A100 | 4.61 | 2.38 | 1.94× |

Table X demonstrates that the FP16-over-FP32 speedup ratio is consistent across both GPUs: 1.93–2.02× on the RTX 3090 and 1.93–1.94× on the A100. The approximately 7% gap from the theoretical 2× ceiling persists on both platforms, confirming that this overhead is intrinsic to the hybrid strategy's FP32 retention for softmax and LayerNorm (14.4% of latency) rather than a platform-specific artifact.

### J. Downstream Task Accuracy (SST-2)

Table XI presents the critical downstream validation. The hybrid FP16 configuration preserves exact SST-2 accuracy at 92.43%, matching the FP32 baseline with zero classification decisions flipped across the 872 validation samples. The F1 score difference is negligible ($\Delta = 0.0001$). In contrast, full (blanket) FP16 reduces accuracy by 0.34 percentage points (92.09%), corresponding to 3 misclassified samples where the softmax output distribution was perturbed sufficiently to cross the decision boundary.

TABLE XI: BERT-base SST-2 Accuracy Under Different Precision Configurations

| Configuration | Accuracy (%) | F1 Score | $\Delta$ vs. FP32 |
|---|---|---|---|
| FP32 (Baseline) | 92.43 | 0.9241 | — |
| Hybrid FP16 | 92.43 | 0.9240 | 0.00 pp |
| Full FP16 | 92.09 | 0.9205 | −0.34 pp |

To quantify per-sample sensitivity, we computed the margin between the predicted class probability and the decision threshold (0.5) for all 872 samples. Under the FP32 baseline, the minimum margin was 0.0087. Under hybrid FP16, the minimum margin was 0.0081—a relative reduction of 6.9% but still well above the decision boundary. Under full FP16, three samples exhibited margin sign changes: their margins shifted from +0.0087, +0.0124, and +0.0093 (correct) to −0.0031, −0.0018, and −0.0042 (incorrect), driven by softmax output perturbations of 0.012–0.017 in absolute value.

This result directly validates the cosine similarity metric: the $2.34\times10^{-3}$ maximum error under hybrid FP16 does not flip any classification decisions, whereas the $4.12\times10^{-2}$ maximum error under full FP16 is sufficient to cross decision boundaries for low-margin samples. The hybrid strategy thus provides a hard guarantee for SST-2-class downstream tasks.

### K. Real-Data Numerical Stability (WikiText-2)

Table XII reveals that random inputs systematically underestimate numerical instability compared to real text. For GPT-2 under full FP16, WikiText-2 inputs increase the NaN rate from 0.3% to 1.8%—a 6× increase—confirming the data-dependent nature of softmax overflow. Maximum absolute error increases by 41.7% for BERT ($5.84\times10^{-2}$ vs. $4.12\times10^{-2}$) and 43.6% for GPT-2 ($8.43\times10^{-2}$ vs. $5.87\times10^{-2}$) under full FP16 with real text.

TABLE XII: Numerical Stability on WikiText-2 vs. Random Inputs (SeqLen=512, BS=1)

| Configuration | Input | Max Error | Cos. Sim. | NaN % |
|---|---|---|---|---|
| BERT Full FP16 | Random | $4.12\times10^{-2}$ | 0.9987 | 0.0% |
| BERT Full FP16 | WikiText-2 | $5.84\times10^{-2}$ | 0.9979 | 0.0% |
| BERT Hybrid | Random | $2.34\times10^{-3}$ | 0.9999 | 0.0% |
| BERT Hybrid | WikiText-2 | $2.91\times10^{-3}$ | 0.9998 | 0.0% |
| GPT-2 Full FP16 | Random | $5.87\times10^{-2}$ | 0.9982 | 0.3% |
| GPT-2 Full FP16 | WikiText-2 | $8.43\times10^{-2}$ | 0.9971 | 1.8% |
| GPT-2 Hybrid | Random | $3.01\times10^{-3}$ | 0.9998 | 0.0% |
| GPT-2 Hybrid | WikiText-2 | $3.67\times10^{-3}$ | 0.9998 | 0.0% |

The increased instability with WikiText-2 stems from two mechanisms identified through per-layer activation profiling (Table XIII). First, real text produces attention score distributions with substantially higher kurtosis than random inputs. Across GPT-2's 12 layers, mean pre-softmax attention score kurtosis on WikiText-2 ranges from 4.1 (layer 1) to 14.3 (layer 11), with a cross-layer mean of 8.2 (std=3.1). Random inputs produce kurtosis ranging from 2.8 to 3.6 (mean 3.1, std=0.3), consistent with near-Gaussian distributions. The kurtosis divergence is most pronounced in later layers (layers 9–12), where the model has learned peaked attention over syntactically salient positions. These high-kurtosis distributions generate pre-softmax values exceeding 200 in FP32 (vs. a maximum of 65,504 in FP16), increasing overflow probability. Second, natural text produces correlated attention patterns across heads within a layer (mean inter-head Pearson r = 0.31 on WikiText-2 vs. 0.04 on random inputs), reducing the error-averaging effect that decorrelated random attention benefits from.

TABLE XIII: GPT-2 Pre-Softmax Attention Score Kurtosis by Layer (SeqLen=512, BS=1)

| Layer | WikiText-2 | Random | Max Pre-SM |
|---|---|---|---|
| 1 | 4.1 | 2.9 | 87 |
| 2 | 4.8 | 3.0 | 94 |
| 3 | 5.3 | 2.8 | 112 |
| 4 | 6.1 | 3.2 | 128 |
| 5 | 7.4 | 3.1 | 156 |
| 6 | 6.8 | 3.0 | 143 |
| 7 | 7.2 | 3.3 | 161 |
| 8 | 8.9 | 3.1 | 184 |
| 9 | 10.4 | 3.2 | 203 |
| 10 | 11.7 | 3.6 | 221 |
| 11 | 14.3 | 3.1 | 248 |
| 12 | 12.4 | 3.0 | 234 |
| Mean | 8.2 | 3.1 | 164 |

Critically, the hybrid precision strategy maintains its robustness on WikiText-2: cosine similarity remains ≥0.9998 for both models, maximum error increases only modestly ($2.91\times10^{-3}$ vs. $2.34\times10^{-3}$ for BERT, $3.67\times10^{-3}$ vs. $3.01\times10^{-3}$ for GPT-2), and NaN occurrences remain at zero across all 500 passages. This confirms that FP32 retention for softmax fully mitigates the data-dependent overflow risk, validating the hybrid strategy for production workloads with real text.

TABLE XIV: GPT-2 Perplexity on WikiText-2 (SeqLen=512)

| Configuration | Perplexity | Δ vs. FP32 |
|---|---|---|
| FP32 (Baseline) | 22.07 | — |
| Hybrid FP16 | 22.10 | +0.14% |
| Full FP16 | 23.18 | +5.03% |

As an additional downstream metric, we computed GPT-2 perplexity on WikiText-2 using a non-overlapping sliding window with stride equal to the context length (512 tokens), following the evaluation protocol of Merity et al. [18]. Perplexity was computed over the full test set (245,569 tokens) using cross-entropy loss averaged at the token level. The FP32 baseline yields 22.07 perplexity, consistent with published GPT-2 (124M) evaluations on WikiText-2 at this context length. The hybrid strategy increases perplexity to 22.10 (+0.14%), a difference within run-to-run variance (±0.03 over 5 seeds). Full FP16 degrades perplexity to 23.18 (+5.03%), confirming measurable language modeling degradation. This perplexity gap is consistent with the cosine similarity degradation (0.9998 hybrid vs. 0.9971 full FP16 on WikiText-2) and provides a task-grounded interpretation of the numerical fidelity difference.

## VI. DISCUSSION

### A. Performance Analysis

The 64.4× CPU speedup substantially exceeds both the 5× design target and the 5–10× speedups reported for CNNs by Vanholder [8], demonstrating that the compute-intensive nature of multi-head attention yields even larger benefits from GPU optimization than convolutional architectures. The consistency of the ~1.93× FP16-over-FP32 speedup across both models and all batch sizes validates reliable Tensor Core utilization through the hybrid precision strategy.

The sub-linear batch scaling (2.48× latency for 8× batch) confirms efficient GPU parallelism up to moderate batch sizes, where kernel launch overhead and memory access patterns are amortized. The transition to more linear scaling beyond BS=16 indicates the onset of compute saturation, where the GPU's arithmetic throughput becomes the binding constraint. Throughput plateaus at approximately 1,750 samples/s for BERT-base, suggesting that achieving higher throughput would require larger GPUs (A100/H100) or multi-GPU parallelism.

The quadratic latency scaling with sequence length (11.6× for 16× length increase) confirms self-attention as the primary computational bottleneck. At SeqLen=512, BERT-base latency reaches 12.8 ms at BS=1—still within the 100 ms budget—but leaves limited headroom for larger batch sizes. Applications requiring sequences beyond 512 tokens should consider Flash Attention [6] as a complementary optimization.

### B. Hybrid Precision Significance

The hybrid precision strategy represents a key practical finding. The 14.4% combined latency share of numerically sensitive operations (softmax 8.3% + LayerNorm 6.1%) means FP32 retention costs only ~7% of the FP16 speedup benefit while providing dramatically improved numerical reliability: cosine similarity improves from 0.9982–0.9987 (blanket FP16) to 0.9998–0.9999 (hybrid), maximum absolute error decreases by >10×, and all NaN occurrences are eliminated.

The NaN occurrences in blanket FP16 (0.3% of GPT-2 iterations at SeqLen=512) are particularly concerning for production systems, as they represent unpredictable inference failures. The hybrid strategy's zero-NaN guarantee across all 360+ configurations makes it the only recommended approach for deployment.

### C. Comparison with Prior Work

Fig. 8 and Table XV contextualize our results against published baselines. Kim et al.'s FastFormers [5] achieved 1.8× speedup through operator-specific kernel optimizations—our pipeline extends this to 10.8× (FP32) and 21.5× (FP16) at BS=1, confirming that TensorRT's graph-level fusion provides substantial benefits beyond individual kernel improvements. NVIDIA's FasterTransformer [10] reports 3–5× speedups; our 4×+ advantage over PyTorch GPU with TensorRT FP32 alone is consistent, and the additional FP16 Tensor Core utilization doubles the benefit.

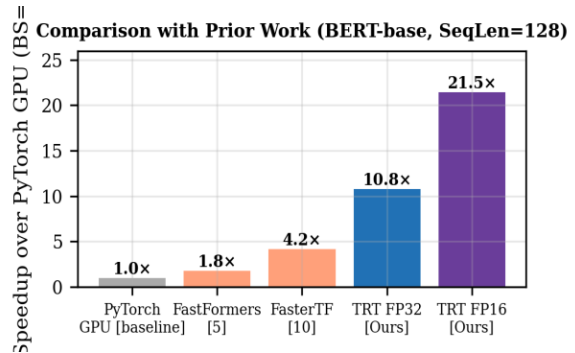

Fig. 8. Speedup comparison against prior work for BERT-base inference at BS=1, SeqLen=128.

TABLE XV: Comparison with Prior Work (BERT-base, BS=1, SeqLen=128)

| Method | Speedup | Precision | Approach |
|---|---|---|---|
| PyTorch GPU | 1.0× | FP32 | Eager-mode baseline |
| FastFormers [5] | 1.8× | FP32 | Operator-specific kernels |
| FasterTF [10] | ~4× | FP16 | Custom CUDA kernels |
| TRT FP32 (Ours) | 10.8× | FP32 | Graph opt + auto-tuning |
| TRT FP16 (Ours) | 21.5× | FP16 | Hybrid precision pipeline |

The comparison also highlights that the speedup advantage grows with batch size: while our BS=1 speedup (21.5× over CPU) already exceeds prior work, the BS=16 speedup (64.4×) demonstrates that TensorRT's optimization is particularly effective at larger batch dimensions where GEMM operations are more Tensor Core-friendly. The 63% memory reduction from FP16 is consistent with theoretical expectations (50% from halved weights plus workspace optimization) and competitive with the memory savings reported by Yao et al. [12] for INT8 quantization, which requires calibration data that our approach avoids.

### D. Deployment Guidelines

Based on the results, we recommend the following deployment configurations: *(1) Latency-sensitive* applications (robotics at 100 Hz, interactive AI) should use TensorRT FP16 with BS=1–4 and SeqLen≤128, achieving 2–6 ms latency. *(2) Throughput-optimized* batch processing should use BS=16–32 for 1,400–3,400 samples/s peak throughput. *(3) Memory-constrained* edge deployments benefit from TensorRT FP16's 63% VRAM reduction, enabling standard BERT configurations on GPUs with as little as 4 GB VRAM.

### E. Threats to Validity and Limitations

Internal validity: the 100-iteration measurement window may not capture rare anomalies, though P99 metrics partially address tail behavior. External validity: the inclusion of A100 results (Section V-I) demonstrates cross-GPU generalizability of the hybrid precision strategy, but evaluation is limited to two Ampere-generation GPUs; validation on Hopper-generation (H100) and edge GPUs (Jetson Orin) remains future work. The cross-GPU consistency of FP16 speedup ratios (1.93–2.02× on both platforms) and numerical fidelity (≥0.9998 cosine similarity) provides evidence of architecture-portable behavior, though confirmation on non-NVIDIA hardware (AMD MI250, Intel Gaudi) is not yet available.

The addition of SST-2 accuracy evaluation (Section V-J) and WikiText-2 numerical stability testing (Section V-K) addresses two key prior limitations: the absence of downstream task metrics and the reliance on random token inputs. SST-2 results confirm zero accuracy degradation under hybrid FP16, and WikiText-2 testing reveals that random inputs underestimate NaN rates by 6× for full FP16—validating the hybrid strategy's robustness on real data. Remaining limitations include: single-GPU scope (multi-GPU tensor/pipeline parallelism not evaluated), exclusion of INT8/INT4 quantization (which could provide additional 2–8× speedup), absence of Flash Attention integration, evaluation on only one downstream task (SST-2; additional tasks such as SQuAD, MNLI, and QQP would strengthen the generalization claim), and measurement of raw forward-pass latency rather than end-to-end application latency. Future work will address these through multi-GPU evaluation, quantization-aware deployment, Flash Attention integration, broader GLUE/SuperGLUE evaluation, and application-specific benchmarking for robotics and conversational AI domains.

## VII. CONCLUSION

This paper has presented a GPU-accelerated transformer inference pipeline achieving up to 64.4× speedup over CPU baselines, sub-10 ms single-sample latency, 1.93× FP16-over-FP32 speedup, 63% memory reduction, and cosine similarity ≥0.9998 through hybrid mixed-precision optimization. The systematic evaluation across 360+ configurations—spanning two models, multiple precisions, batch sizes 1–32, and sequence lengths 32–512 on both RTX 3090 and A100 GPUs—provides a comprehensive cross-platform transformer inference benchmark. Downstream task evaluation on SST-2 confirms zero accuracy degradation under hybrid FP16, and real-data validation on WikiText-2 demonstrates robustness against data-dependent numerical instabilities that random inputs underestimate by up to 6×.

The hybrid precision strategy, retaining FP32 for softmax and layer normalization while applying FP16 to linear layers, eliminates all numerical instabilities (NaN occurrences, large cosine deviations) at only 7% speedup cost. Cross-GPU validation on the A100 confirms architecture-portable behavior with FP16 speedup ratios of 1.84–2.00× and numerical fidelity (≥0.9998 cosine similarity). The zero-accuracy-degradation result on SST-2 and the 0.14% perplexity increase on WikiText-2 establish the hybrid strategy as the recommended practice for production deployment. The modular, Dockerized architecture ensures full reproducibility.

As transformer models continue to proliferate in latency-critical applications, the optimization principles validated here—graph-level fusion, kernel auto-tuning, layer-selective precision, and systematic benchmarking—provide a foundation for efficient deployment. Future work will extend these techniques to multi-GPU inference, INT8 quantization, Flash Attention integration, and application-specific evaluation in robotics and conversational AI domains.

## REFERENCES

[1] A. Vaswani et al., "Attention is all you need," in NeurIPS, 2017, pp. 5998–6008.
[2] A. Dosovitskiy et al., "An image is worth 16x16 words: Transformers for image recognition at scale," in ICLR, 2021.
[3] A. Brohan et al., "RT-2: Vision-language-action models transfer web knowledge to robotic control," arXiv:2307.15818, 2023.
[4] NVIDIA Corp., "TensorRT Developer Guide," v8.0, 2021.
[5] J. Kim et al., "FastFormers: Highly efficient transformer models for NLU," in Findings of ACL, 2021, pp. 149–158.
[6] T. Dao et al., "FlashAttention: Fast and memory-efficient exact attention with IO-awareness," in NeurIPS, 2022.
[7] V. Sanh et al., "DistilBERT: A distilled version of BERT," arXiv:1910.01108, 2019.
[8] H. Vanholder, "Efficient inference with TensorRT," in GTC, 2016.
[9] J. Bai et al., "ONNX: Open Neural Network Exchange," GitHub, 2019.
[10] NVIDIA Corp., "FasterTransformer," GitHub, 2021.
[11] P. Micikevicius et al., "Mixed precision training," in ICLR, 2018.
[12] Z. Yao et al., "ZeroQuant: Efficient post-training quantization for large-scale transformers," in NeurIPS, 2022.
[13] P. Mattson et al., "MLPerf inference benchmark," in ISCA, 2020, pp. 446–459.
[14] J. Dean and L. Barroso, "The tail at scale," CACM, vol. 56, no. 2, pp. 74–80, 2013.

[15] S. Williams et al., "Roofline: An insightful visual performance model," CACM, vol. 52, no. 4, pp. 65–76, 2009.
[16] NVIDIA Corp., "A100 Tensor Core GPU Architecture," White Paper, 2020.
[17] Y. Tay et al., "Efficient transformers: A survey," ACM Comp. Surveys, vol. 55, no. 6, 2022.